\documentclass[10pt,twocolumn,letterpaper]{article}

\usepackage{abcd}
\usepackage{enumitem}
\usepackage{subcaption}
\usepackage{dsfont}
\usepackage{pifont}
\usepackage{blindtext}
\usepackage{morewrites}

\usepackage{times}
\usepackage{epsfig}
\usepackage{graphicx}
\usepackage{amsmath}
\usepackage{amssymb}
\usepackage{float}
\restylefloat{table}

\usepackage{multirow}
\usepackage{comment}
\newcommand{\xmark}{\ding{55}}
\newcommand{\cmark}{\ding{51}}
\usepackage{makecell}
\makeatletter

\usepackage[pagebackref=true,breaklinks=true,letterpaper=true,colorlinks,bookmarks=false]{hyperref}

\abcdfinalcopy

\ifabcdfinal\fi
\begin{document}

\title{Beyond Domain Adaptation: Unseen Domain Encapsulation via \\ Universal Non-volume Preserving Models}

\author{Thanh-Dat Truong $^{1}$, Chi Nhan Duong $^{2}$, Khoa Luu $^{3}$, Minh-Triet Tran $^{1}$, Minh Do  $^{4}$\\
	$^{1}$ University of Science, Vietnam\\
	$^{2}$ Computer Science and Software Engineering, Concordia University, Canada\\
	$^{3}$ Computer Science and Computer Engineering, University of Arkansas, USA\\
	$^{4}$ University of Illinois, USA\\
	\tt\small ttdat@selab.hcmus.edu.vn, dcnhan@ieee.org, khoaluu@uark.edu, \\ \tt\small tmtriet@fit.hcmus.edu.vn,   minhdo@illinois.edu 
}

\maketitle

\begin{abstract}

Recognition across domains has recently become an active topic in the research community. However, it has been largely overlooked in the problem of recognition in new unseen domains. Under this condition, the delivered deep network models are unable to be updated, adapted or fine-tuned. Therefore, recent deep learning techniques, such as: domain adaptation, feature transferring, and fine-tuning, cannot be applied. This paper presents a novel Universal Non-volume Preserving approach to the problem of domain generalization in the context of deep learning. The proposed method can be easily incorporated with any other ConvNet framework within an end-to-end deep network design to improve the performance. On digit recognition, we benchmark on four popular digit recognition databases, i.e.  MNIST, USPS, SVHN and MNIST-M. The proposed method is also experimented on face recognition on Extended Yale-B, CMU-PIE and CMU-MPIE databases and compared against other the state-of-the-art methods. In the problem of pedestrian detection, we empirically observe that the proposed method learns models that improve performance across a priori unknown data distributions. 
\end{abstract}

 \begin{figure}[t]
	\centering \includegraphics[width=1.0\columnwidth]{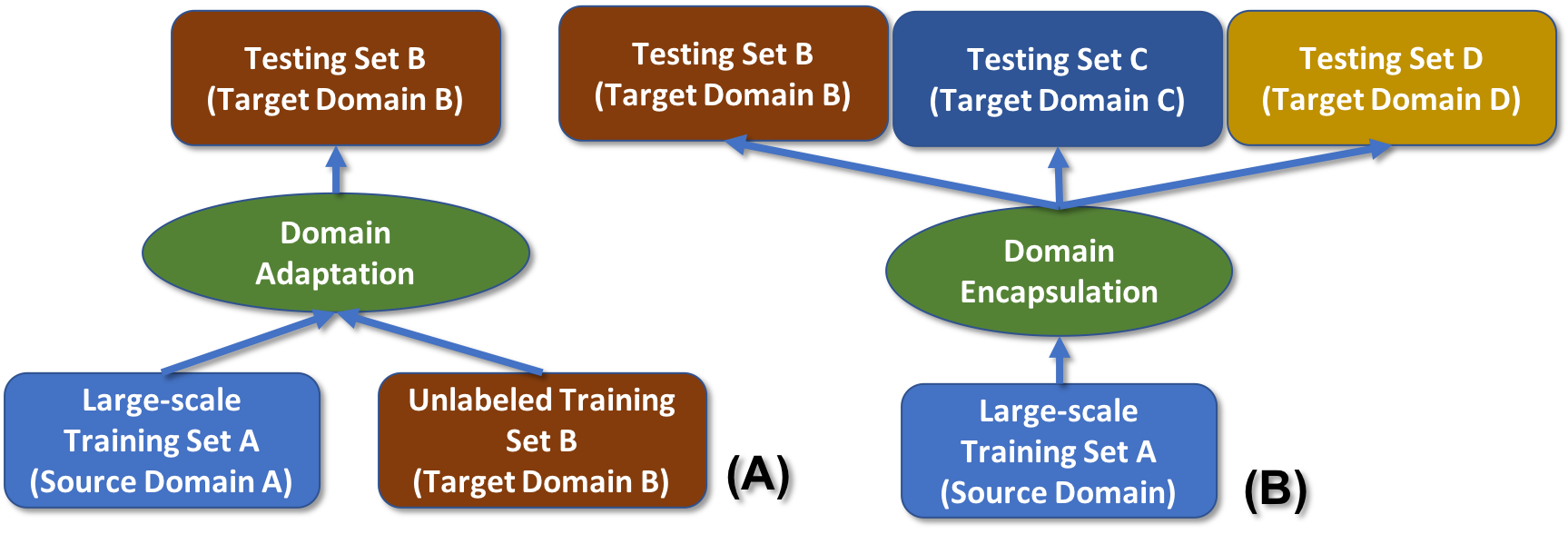}
	\caption{Comparison between Domain Adaptation (A) and our proposed Domain Encapsulation (B) problems}	
	\label{fig:Fig1}
\end{figure}

\section{Introduction}

Deep learning-based detection  and recognition studies have been recently achieving very accurate performance in visual applications. However, many such methods expect the test images to come from the same distribution as the training images, and often fail when presented with new unseen visual domains. For examples, in face recognition, a system is trained on RGB images/videos and then deployed on infrared or thermal images/videos. Far apart from domain adaptation \cite{pmlr-v37-ganin15, adda_cvpr2017}, feature transfer learning \cite{feature_transfer_learning} or feature fine-tuning \cite{finetune_learning}, in the context of the proposed problem, there is no available information about new unseen environments or domains where the system will be deployed.
 
\begin{figure*}[t]
	\centering \includegraphics[width=1.99\columnwidth]{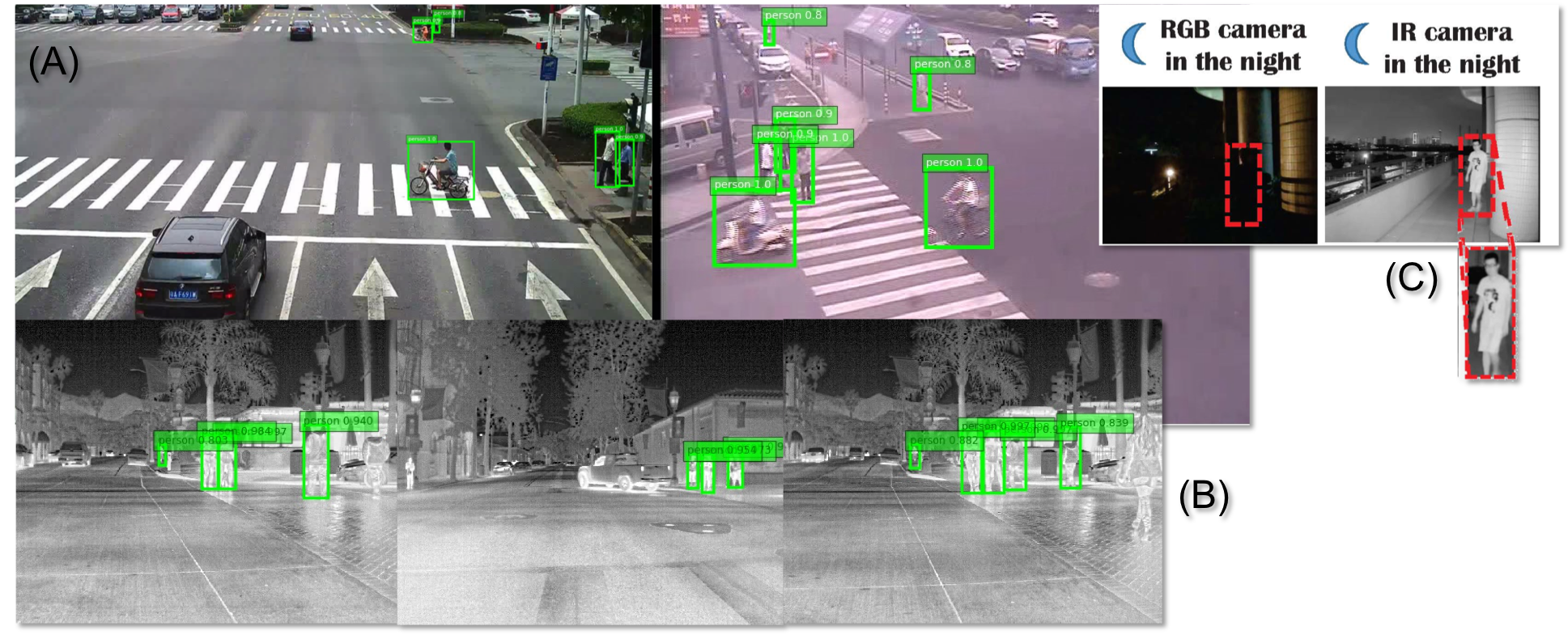}
	\caption{The ideas of unseen domain encapsulation. The deep model is trained only in a single domain (A), i.e. RGB images. It is deployed in other unseen domains, i.e. thermal (B) and infrared (C) images}	
	\label{fig:test_domains}
\end{figure*} 
 
Indeed, detection and classification crossing domains have recently become active topics in the research communities. In particular, domain adaptation has received significant attention in computer vision. As shown in Figure \ref{fig:Fig1}(A), in the domain adaptation, we usually have a large-scale training set with groundtruth, i.e. the source domain A, and a small training set with or without groundtruth, i.e. the target domain B. The knowledge from the source domain A will be learned and adapted to the target domain B. During the testing time, the trained model will be deployed \textit{only} in the target domain B. The recent results in domain adaptation have showed significant improvement in the performance. However, in real-world applications, the trained models are potentially deployed not only in the target domain B but also in many other \textit{new unseen domains}, e.g. C, D, etc. In this scenarios, the released deep network models are usually unable to be retrained or fine-tuned with the inputs in new domains or environments as shown in Figure \ref{fig:test_domains}. Thus, domain adaptation methods cannot be applied in these problems since the new unseen target domains are unavailable during training. Moreover, domain adaptation methods only allow a pair of domains, i.e. the source domain and the target domain. Meanwhile,  real-world applications usually require more than just a pair of domains. In practice, the number of domains that released models are potentially deployed is usually large and unpredictable.

Besides, there are some prior work to perform recognition problems with high accuracy by presenting new loss functions \cite{triplet_loss, range_loss} or increasing deep network structures \cite{Huang_2017_CVPR} via mining hard samples in training sets. These loss functions are deployed to deal with hard samples that can be considered as unseen domains, one can consider it as hard-sample problems and then can be solved using new loss functions, e.g. Center Loss \cite{center_loss}, Range Loss \cite{range_loss}, etc. However, these methods are also limited to generalize in new unseen domains.
Increasing the depth of the deep network can be also considered to deal with hard samples. However, some real world problems are unable to observe the training samples in new unseen domains during the training process.
Therefore, in the scope of this work, there is no assumption about the new unseen domains. 
Our proposed method can be supportively incorporated with these Convolutional Neural Network (CNN) based detection and classification methods to train within end-to-end deep learning framework to potentially improve the performance.

\begin{table*} [!t]
\small
\centering
\caption{Comparing the properties between our UNVP approach and other recent methods. Feature Transferring (FT), Adversarial Discriminative Domain Adaptation (ADDA), Domain-Adversarial Training of Neural Networks (DANN), Generalizing to Unseen Domain via Adversarial Domain Adaptation(ADA), Universal Background Models (UBM), Coupled Generative Adversarial Networks (CoGAN), DL (Deep Learning), Generative Adversarial Network (GAN), Convolutional Neural Network (CNN), Adversarial Loss ($l_{adv}$), Log Likelihood Loss (LL), Cycle Consistency Loss ($l_{cyc}$)}
	
\begin{tabular}{| p{2.5cm}   |  p{1.6cm}     | p {1.1cm}     | p {1.4cm}           | p {1.1cm}          | p {1.4cm}          |p {1.7cm}            | p {1.1cm}         | p {1.15cm}|}
	\hline
	                        &  \textbf{Our UNVP} & FT\cite{feature_transfer_learning} & ADDA\cite{adda_cvpr2017} & DANN\cite{pmlr-v37-ganin15} & CoGAN\cite{NIPS2016_6544} & I2IAdapt\cite{Murez_2018_CVPR}& ADA\cite{generalize-unseen-domain} & UBM\cite{Reynolds00speakerverification} \\ 
	\hline \hline
	
	\textbf{Model Type}     & \textbf{DL}            & DL            & DL                  & DL                 & DL                 & DL                    & DL                & GMM \\
	\hline
	
	\textbf{Architecture}   & \textbf{PGM+CNN}       & CNN           & CNN+GAN             & CNN                & CNN+GAN            &  CNN+GAN              & CNN               & PGM \\
	\hline
	
	\textbf{Loss Function}  & \textbf{LL+$\ell_2$}   & $\ell_2$      & $l_{adv}$           & $l_{adv}$          & $l_{adv}$          & $l_{adv}$+$\l_{cyc}$    & $\ell_2$        & \xmark LL \\
	\hline
	
	\textbf{End-to-End}     & \textbf{Yes}           &\cmark Yes     & \cmark Yes          & Yes                & Yes                &\cmark    Yes          & \xmark Yes        & \xmark No \\
	\hline
	
	\textbf{Require samples in new domains} 
	                        & \textbf{No}            &\cmark Yes     & \cmark Yes          & Yes                & Yes                &\cmark Yes             & \xmark No         & \xmark Yes \\
	\hline
	
    \textbf{Pair Domains}   & \textbf{Any}           & Two           & Two                 & Two                & Any                & Two                   & \xmark Any        & \xmark Any \\
    	\hline
\end{tabular}\label{tab:TenMethodSumm}
\end{table*}

In this paper, instead of the domain shift, we target on exploring the problem of domain generalization in the context of deep learning.
This work is inspired from the Universal Background Models (UBM) \cite{Reynolds00speakerverification} to model the background environment in the speech verification system. However, instead of approaching the background or environment modeling by using Gaussian Mixture Models (GMMs), this work  presents a novel approach to generalize the domain representation by a new deep network design. 

\subsection{Contributions of this Work}

This work presents a novel approach to domain encapsulation that can learn to better generalize new unseen domains. The restrictive setting is considered in this work where there is only single source domain for training data. 
Table \ref{tab:TenMethodSumm} summarizes the difference between our approach and the prior methods.
The contribution of this work can be summarized as follows.

A novel approach named \textit{Universal Non-volume Preserving Models (UNVP)} is firstly introduced to generalize environments of new unseen domains from a given single source training domain. Secondly, the environmental features extracted from the environment modeling via UNVP and the discriminative features extracted from the deep network classifiers are then unified together to provide a final encapsulated deep features that are robustly discriminative in new unseen domains. The proposed UNVP approach is designed and implemented within an end-to-end deep learning framework and inherits the power of the Convolutional Neural Network.
The proposed UNVP can be easily end-to-end integrated with a CNN deep network design for object detection, object recognition or segmentation so that it can perform improvement results.
Finally, the proposed method will be experimented in numerous vision modalities and applications with improvement results to demonstrate the impact of the method.

\section{Related work}

This section first overviews the Universal Background Models method. Then, the recent work in domain adaptation and image-to-image translation are also summarized.

\subsection{Universal Background Models (UBM)} 

A Universal Background Models is a destiny estimation modeling method originally used in the speaker verification system \cite{Reynolds00speakerverification}.
In the conventional Gaussian Mixture Model - Universal Background Models (GMM-UBM) framework, UBM is a GMMs trained on a pool of data, known as the background or the environment from a large number of speakers.
The speaker-specific models are then adapted from the UBM using the Maximum a Posteriori Probability (MAP) parameter estimation.
During the evaluation phase, each test segment is scored against all enrolled speaker models to determine the speaker identities, i.e. speaker identification. It is also scored against the background model and a given speaker model to accept or reject an identity claim, i.e. speaker verification.

\subsection{Domain Adaptation}

Domain adaptation has recently become one of the most popular research topics in the field \cite{pmlr-v37-ganin15, DBLP:journals/corr/TzengHDS15, Sener:2016:LTR:3157096.3157333, DBLP:journals/corr/TzengHZSD14, adda_cvpr2017}. The key idea of domain adaptation is to map both source and target domains into a common feature space.
Tzeng et al. proposed a unified framework for unsupervised domain Adaptation based on Adversarial learning objectives (ADDA) \cite{adda_cvpr2017}. 
It uses a loss function in the discriminator to be solely dependent on its target distribution. 
Ganin et al. proposed a method for both training and domain adaptation within a unified network \cite{pmlr-v37-ganin15}. It aims to learn both domain adaptation and classification at the same time. 

\subsection{Image-to-Image Translation}

The research Image-to-Image Translation topic has been grown significantly for several years. Image Translation has many potential applications, e.g. image style transfer \cite{Dundar2018DomainSA}, semantic segmentation \cite{pix2pix2017, wang2017highres}, depth estimation \cite{abarghouei18monocular}, etc. 
Indeed, Domain Adaptation is the one of important applications of Image Translation \cite{Murez_2018_CVPR, Bousmalis_2017_CVPR, NIPS2017_6672, huang2018munit}.

Liu et. al. presented Coupled Generative Adversarial Network (CoGAN) for learning a joint distribution of multi-domain images \cite{NIPS2016_6544} which applies for domain adaptation. 
Liu et al. introduced Unsupervised Image to Image Translation (UNIT) method \cite{NIPS2017_6672} inherited from CoGAN. 
This method aims at learning joint distribution of two marginal distributions from two image domains. 
The shared-latent space assumption was used in CoGAN \cite{NIPS2016_6544} for joint distribution learning.
In order to improve UNIT method, Huang et al. presented Multimodal UNIT (MUNIT) \cite{huang2018munit}. This method assumes that the image representation can be decomposed into a content code that is domain-invariant, and a style code that captures domain-specific properties.
It allows images in different domains can be decoded into the shared feature space.

\begin{figure}[!t]
	\centering \includegraphics[width=0.95\columnwidth]{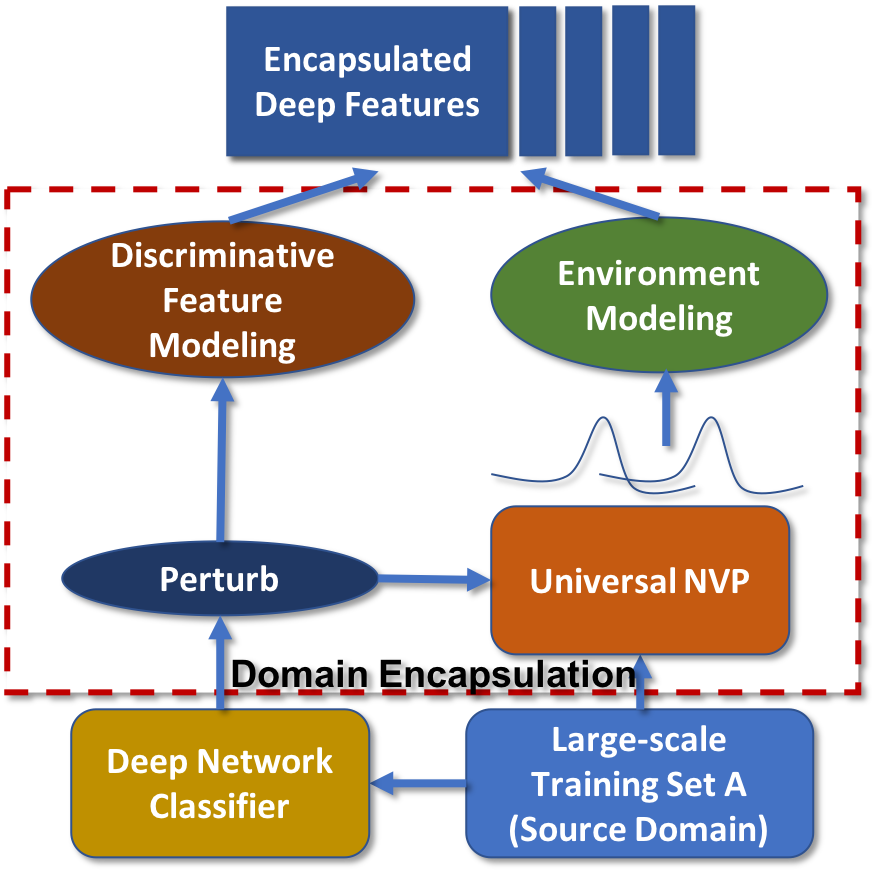}
	\caption{The Proposed Method}
	\label{fig:Method}
\end{figure}

 \begin{figure*}[t]
 	\centering \includegraphics[width=1.99\columnwidth]{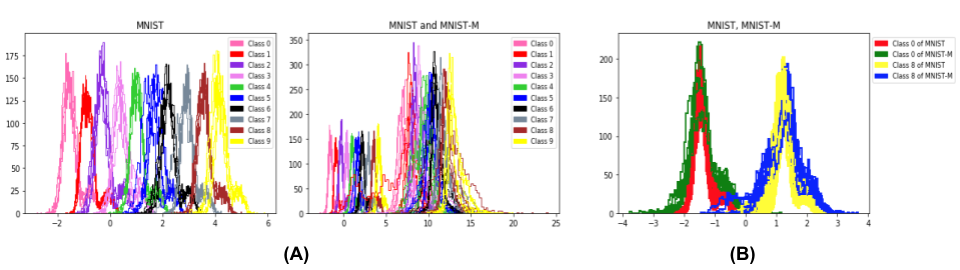}
 	\caption{The distributions: (A) The left figure is the distribution of MNIST dataset. The right figure is the distributions of MNIST and MNIST-M datasets. The right image illustrates that when modeling MNIST-M using the model trained on MNIST, the distribution of MNIST-M is shifted to the right. (B) The distributions of class 0 and 8 of MNIST and MNIST-M after using unseen domain generalization to train environment variation modeling.}
 	\label{fig:Distribution}
 \end{figure*}

\section{The Proposed Universal Non-volume Preserving Models (UNVP)}

The proposed UNVP approach presents a new \textit{tractable CNN deep network} to not only extract the deep CNN features but also formulate the probability densities of the samples in the source environment in the form of Gaussian distributions.  From these learned distributions, a density-based augmentation approach is employed to expand data distribution of the source environment for generalizing to different unseen domains. Far apart from other augmentation techniques where samples are generated directly in image domain using some prior knowledge, e.g. synthesizing blurry version of the image, or adding synthesis backgrounds, our approach focuses on augmentation in semantic space via the estimation of environment density. As a results, more semantic samples are augmented and, therefore, generalize the learning process.
With this architecture design, UNVP is clever to unify the power of CNN deep features modeling in the first phase in the network and the distribution modeling in the later phase in the network within an end-to-end training framework as shown in Figure \ref{fig:Method}. In particular, the proposed framework consists of three main components. (1) \textit{Domain variation UNVP modeling} via deep mapping functions; (2) \textit{Unseen domain generalization}; and (3) \textit{End-to-end joint training deep network}.

\subsection{Environment Variation Modeling via Density Functions}

Modeling environment variation directly in high-dimensional image domain is extremely complicated and easy to diverse due to the effects of noisy samples.
This section aims at learning a function $\mathcal{F}$ that maps an image $\mathbf{x}$ in image domain $\mathcal{I}$ to its latent representation $\mathbf{z}$ in latent domain $\mathcal{Z}$ such that the density function of $\mathbf{x}$ can be estimated via 
the probability density function $p_Z(\mathbf{z})$.
Then via $\mathcal{F}$, rather than representing the environment variation directly in the image domain, it can be easily modeled via variables in latent space that provides more semantic manner. 
\noindent
\textit{\textbf{Structure and Variable Relationship.}}
Let $\mathbf{x} \in \mathcal{I}$ be a data sample in image domain $\mathcal{I}$, $y$ be its corresponding class label, and $\mathbf{z} = \mathcal{F}(\mathbf{x}, y,\theta)$ where $\theta$ denotes the parameters of $\mathcal{F}$, the probability density function of $\mathbf{x}$ can be formulated via the change of variable formula as follows:
\begin{equation} \label{eqn:DensityFunc}
\small
    p_X(\mathbf{x},y;\theta) = p_Z(\mathbf{z},y;\theta)\left|\frac{\partial \mathcal{F} (\mathbf{z},y;\theta)}{\partial\mathbf{x}} \right|
\end{equation}
where $p_X(\mathbf{x},y)$ and $p_Z(\mathbf{z},y;\theta)$ define the distributions of samples of class $y$ in image and latent domains, respectively. $\frac{\partial \mathcal{F} (\mathbf{z}, y;\theta)}{\partial\mathbf{x}}$ denotes the Jacobian matrix with respect to $\mathbf{x}$. Then the log-likelihood is computed by.

\begin{equation} \label{eqn:LogLikelihoodFunc}
\small
    \log p_X(\mathbf{x},y;\theta) = \log p_Z(\mathbf{z},y;\theta) + \log\left|\frac{\partial \mathcal{F} (\mathbf{z},y;\theta)}{\partial\mathbf{x}} \right|
\end{equation}

Eqns. \eqref{eqn:DensityFunc} and \eqref{eqn:LogLikelihoodFunc} have provided two facts: (1) learning the density function of samples in class $y$ is equivalent to estimate the density of its latent representation $\mathbf{z}$ and determinant of the associated Jacobian matrix $\frac{\partial \mathcal{F}}{\partial \mathbf{x}}$; and (2) if the latent distribution $p_Z$ is defined as a Gaussian distribution, the learned function $\mathcal{F}$ explicitly becomes the mapping function from a real data distribution to a Gaussian distribution in latent space. 
Then, we can model the environment variation via deviations from the Gaussian distributions of all classes in latent domain.
Furthermore, when $\mathcal{F}$ is well-defined with tractable computation of its Jacobian determinant, the two-way connection (i.e. inference and generation) can be established between $\mathbf{x}$ and $\mathbf{z}$. 

\noindent
\textbf{\textit{The prior class distributions.}} Motivated from these properties, given $C$ classes, we choose the Gaussian distributions with different means $\{\boldsymbol{\mu}_1, \boldsymbol{\mu}_2, .., \boldsymbol{\mu}_C\}$ and covariances $\{\Sigma_1,\Sigma_2,...,\Sigma_C\}$ as prior distributions for these classes, i.e. $\mathbf{z}_c \sim \mathcal{N}(\boldsymbol{\mu}_c, \Sigma_c)$. 

\noindent
\textit{\textbf{Mapping Function Structure.}}
In order to enforce the information flow from image domain to latent space with different abstraction levels, the mapping function $\mathcal{F}$ is formulated as a composition of several sub-functions $f_i$ as follows.
\begin{equation}
\small
    \mathcal{F} = f_1 \circ f_2 \circ ... \circ f_N
\end{equation}
where $N$ is the number of sub-functions. The Jacobian $\frac{\partial \mathcal{F}}{\partial \mathbf{x}}$ can be derived by $\frac{\partial \mathcal{F}}{\partial \mathbf{x}} = \frac{\partial f_1}{\partial \mathbf{x}} \cdot \frac{\partial f_2}{ \partial f_1} \cdots \frac{\partial f_N}{ \partial f_{N-1}}$. With this structure, the properties of each $f_i$ will define the properties for the whole mapping function $\mathcal{F}$. For example, if the Jacobian of $\frac{\partial f_1}{\partial \mathbf{x}}$ is tractable, then $\mathcal{F}$ is also tractable. Furthermore, if $f_i$ is a non-linear function built from a composition of CNN layers then $\mathcal{F}$ becomes a deep convolution neural network. There are several ways to construct the sub-functions, i.e. borrowing different CNN structures for non-linearity property. In our approach, the sub-function in \cite{Duong_2017_ICCV} is adopt thanks to its tractable and invertible.
\begin{equation} \nonumber
\small
    f(\mathbf{x}) = \mathbf{b} \odot \mathbf{x} + (1-\mathbf{b) \odot \left[\mathbf{x} \odot \exp{(\mathcal{S}(\mathbf{b} \odot \mathbf{x})} + \mathcal{T}(\mathbf{b} \odot \mathbf{x})\right]}
\end{equation}
where $\mathbf{b}$ is a binary mask, and $\odot$ is the Hadamard product. $\mathcal{S}$ and $\mathcal{T}$ define the scale and translation functions during mapping process.

\noindent
\textit{\textbf{Learning the mapping function and Environment Modeling.}}
In order to learn the parameter $\theta$ for mapping function $\mathcal{F}$, the log-likelihood in Eqn. \eqref{eqn:LogLikelihoodFunc} is maximized as follows.
\begin{equation}
    \theta^* = \arg \max_{\theta} \sum_{c}\sum_{i}\log p_X(\mathbf{x}^i,c;\theta)
\end{equation}

Notice that after learning the mapping function, all images of all classes are mapped into the distributions of their classes. Then the environment density can be considered as the composition of these distributions. Figure \ref{fig:Distribution}(A) (left) illustrated an example of the learned environment distributions of MNIST dataset with 10 digit classes. The density distributions of two different environments (i.e. MNIST and MNIST-M) are also presented in Figure \ref{fig:Distribution}(A) (right). In the next section, a generalization approach is proposed so that using only samples in source environment, the learned model can expand the density distributions of source environment so that they can cover as much as possible the distributions of unseen environments.

\subsection{Unseen Domain Generalization} \label{sec:DomainGeneralization}

After modeling the source environment variation as the compositions of its class distributions, this section introduces the generalization process of these distributions with respect to a classification model $\mathcal{M}$ such that the expansion of these distributions can help $\mathcal{M}$ generalize to unseen environments with high accuracy.

In particular, let $\ell(\mathbf{X,Y};\mathcal{M},\theta, \theta_1)$ be the training loss function of $\mathcal{M}$, and $\theta_1$ be the parameters of $\mathcal{M}$. The generalization process of $\mathcal{M}$ can be formulated as updating the parameters $\theta_1$ such that even the class distributions of the unseen environment are distance $\rho$ away from the source environment, $\mathcal{M}$ is still robust with high accuracy.
The objective function is reformulated as.
\begin{equation} \label{Eqn:WorseCaseFormulation}
    \arg \min_{\theta_1} \sup_{P:d(P_X,P_X^{src})\leq \rho} \mathbb{E}\left[ \ell(\mathbf{X,Y};\mathcal{M},\theta, \theta_1)\right] 
\end{equation}
where $\{\mathbf{X,Y}\}$ denotes the sets of images and their labels; $d(\cdot, \cdot)$ is the distance between probability distributions; $P^{src}_X(\mathbf{X,Y})$ and $P_X(\mathbf{X,Y})$ be the density distributions of the source and the new unseen environments, respectively.

Since both $P_X^{src}$ and $P_X$ are density distributions, the Wasserstein distance with respect to $P_X^{src}$ and $P_X$ can be adopt as follows.
\begin{equation}
    d(P_X,P_X^{src}) = \sum_{c} \inf \mathbb{E} \left[ cost\left( \mathbf{\bar{X}}^c,\mathbf{X}^c\right) \right]
\end{equation}
where $cost(\cdot, \cdot)$ denotes the transformation cost. Notice that from previous section, we have leaned a mapping function $\mathcal{F}$ that maps the density functions from image space to prior distribution in latent space. Moreover, since $\mathcal{F}$ is invertible with the specific formula of sub-function, computing $d(P_X,P_X^{src})$ is equivalent to $d(P_Z,P_Z^{src})$. From this, we can estimate $cost$ as the transformation cost between Gaussian distributions. 
Then $d(P_X,P_X^{src})$ is reformulated by.
\begin{equation} \label{eqn:DistanceW}
\small 
\begin{split}
    &d^2(P_X,P_X^{src}) = d^2(P_Z,P_Z^{src})\\
    &=\sum_c || \mu{'}_c - \mu_c ||^2_2 
    + \text{Tr}(\Sigma'_c + \Sigma_c - 2(\Sigma'^{1/2}_c\Sigma_c\Sigma'^{1/2}_c)^{1/2})
\end{split}    
\end{equation}
where $\{\mu_c, \Sigma_c\}$ and $\{\mu'_c, \Sigma'_c\}$ are the means and covariances of the distributions of class $c$ in the source environment and unseen environment, respectively.
Plugging this distance and applying the Lagrangian relaxation to Eqn. \eqref{Eqn:WorseCaseFormulation}, we have
\begin{equation}
\small \nonumber
\begin{split}
    &\arg \min_{\theta_1} \sup_P \mathbb{E} \left[\ell(\mathbf{X,Y};\mathcal{M},\theta, \theta_1)\right] - \alpha \cdot d(P_X,P_X^{src})\\
    =&\arg \min_{\theta_1} \sum_c \sup_x \{ \ell(\mathbf{x},c;\mathcal{M},\theta, \theta_1) - \alpha \cdot cost(\mathbf{x},\mathbf{x}_c^{src}) \}
\end{split}
\end{equation}
To solve this objective function, the optimization process can be divided into two alternative steps: (1) generate the sample $\mathbf{x}$ for each class such that $\mathbf{x}=\arg \max_{\mathbf{x}} \{ \ell(\mathbf{x},c;\mathcal{M},\theta, \theta_1) - \alpha \cdot cost(\mathbf{x},\mathbf{x}_c^{src}) \}$, and consider this is a new ``hard'' example for class $c$; and (2) add $\mathbf{x}$ to the training data and optimize the model $\mathcal{M}$. In other words, this two-step optimization process aims at finding new samples belonging to distributions that are $\rho$ distance far away from the distributions of the source environment, and making $\mathcal{M}$ became more robust when classifying these examples. 
By this way, after a certain of iteration, the distributions learned from $\mathcal{M}$ can be generalized so that they can cover as much as possible the distributions of new unseen environments.

Figure \ref{fig:Distribution}(B) shows that the distributions of MNIST and MNIST-M have been joint after using unseen domain generalization to train environment variation modeling. The red and green line of Figure \ref{fig:Distribution}(B) illustrate the distribution of class 0 of MNIST and MNIST-M, respectively. The yellow and blue line of Figure \ref{fig:Distribution}(B) illustrate the distribution of class 8 of MNIST and MNIST-M, respectively. Its figure proves that by our method can cover both distributions of the source domain and distributions of an unseen domain.

\begin{figure}[t]
	\centering \includegraphics[width=1.0\columnwidth]{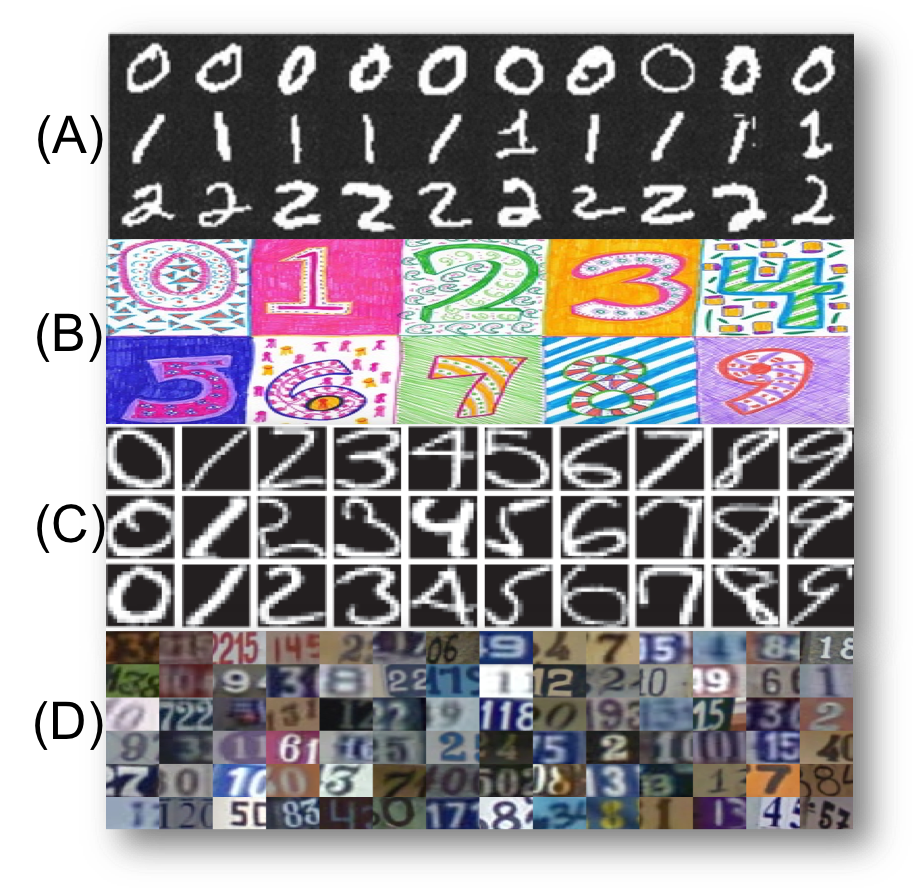}
	\caption{Examples in (A) MNIST, (B) MNIST-M, (C) USPS and (D) SVHN databases}	
	\label{fig:char_dbs}
\end{figure}

\subsection{Universal Deep Models}

The whole end-to-end joint training process for Universal Deep Models is illustrated in Figure \ref{fig:Method}. Given a large-scale training set in the source environment, UNVP is employed to learn the mapping function from image domain to distributions in latent space. 
Then the two-step training process as presented in Sec. \ref{sec:DomainGeneralization} is adopted to train the Deep Classifier $\mathcal{M}$ for generalization. Notice that, to further constraint the perturbation in latent space while generating for new samples $\mathbf{x}$, we incorporate a regularization of the latent space learned by the classifier to Eqn. \eqref{eqn:DistanceW} as follows:
\begin{equation} 
\small \nonumber
\begin{split}
    &d^2(P_X,P_X^{src}) \\
    =&\sum_c || \mu{'}_c - \mu_c ||^2_2 
    + \text{Tr}(\Sigma'_c + \Sigma_c - 2(\Sigma'^{1/2}_c\Sigma_c\Sigma'^{1/2}_c)^{1/2}) \\
    & + || \mathcal{M}(\mathbf{X}_c) - \mathcal{M}(\mathbf{X}'_c)||_2^2
\end{split}    
\end{equation}
New generated samples are then added to the training set and used for updating both UNVP and CNN classifiers. 

\section{Experimental Results}

\begin{figure}[t]
	\centering \includegraphics[width=1.0\columnwidth]{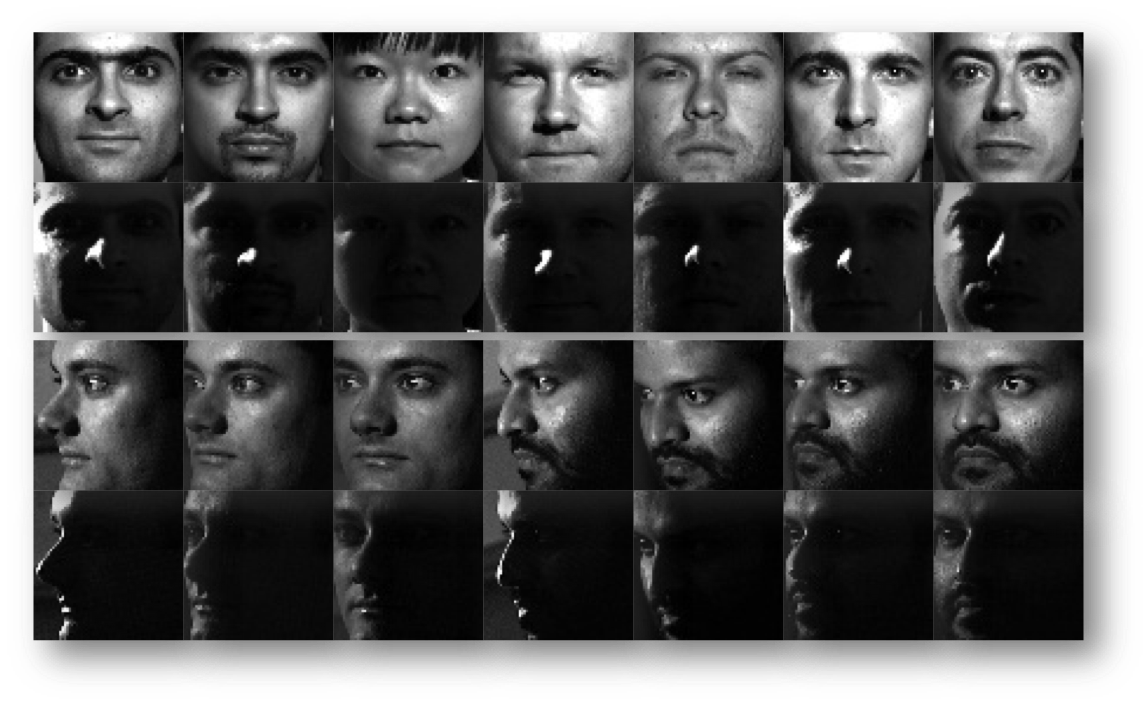}
	\caption{Examples of Yale-B \cite{yale_b_dataset} and CMU-PIE \cite{pie_dataset} databases. Face images in normal illumination condition (the 1st and the 3rd rows) are used as training domain and the ones in dark illumination conditions (the 2nd and the 4th rows) are used for testing as new unseen domains.}	
	\label{fig:face_dbs}
\end{figure}

This section first validates the proposed approach in digit recognition on four digit datasets, i.e.  MNIST \cite{mnist_dataset}, USPS \cite{usps_dataset}, SVHN \cite{svhn_dataset} and MNIST-M. To obtain the MNIST-M, we blend digits from the original set over patches randomly extracted from color photos from BSDS500 \cite{countour_detection}. In this experiment, MNIST is used as the only training set and the others are used as the testing sets.
Then, Subsection \ref{subsec:face} shows the proposed approach in face recognition with three standard face recognition databases, i.e. Extended Yale-B \cite{yale_b_dataset}, CMU-PIE \cite{pie_dataset}, CMU-MPIE \cite{multi_pie_dataset}. Facial images with normal illumination are used as training domain and the ones in dark illumination conditions are used as testing set on the new unseen domains (Figure \ref{fig:face_dbs}).
Finally, we show the advantages of our proposed method in the cross-domain pedestrian recognition, i.e. RGB and thermal domains. we compare the detection results using our proposed method against other standard methods in subsection \ref{subsec:ped}.

\subsection{Digit Recognition on Unseen Domains}
\label{subsec:digit}

We show the experimental results using our proposed approach to digit recognition on new unseen domains with four digit databases, i.e. MNIST, MNIST-M, USPS and SVHN, as shown in Figure \ref{fig:char_dbs}. In order to simplify it, LeNet CNN deep network model \cite{lenet_ref} is used as the classifier in this experiment. This deep network can be technically replaced by any other deep network models. 
We use a ConvNet with the designed architecture \textit{conv-pool-conv-pool-fc-fc-softmax}. For environment variation modeling, we use Real NVP \cite{real_nvp} as domain variation UNVP modeling.

About the training network hyper-parameters, learning rate, batch size and regularization rate are set to $0.0001, 256, 0.00005$, respectively. In the generalizing phase (it is equivalent to \textit{maximization} of ADA), we set hyper-parameters $\eta, T_{min}, T_{max}, K$ to $1.0, 100, 15, 6$, respectively, as shown in \textit{\textbf{ Algorithm 1}} of ADA \cite{generalize-unseen-domain}. Adam Optimizer is used to optimize and update the deep network.

In this experiment, MNIST is the only database used to train the classifier. Then, three other datasets, i.e. MNIST-M, USPS and SVHN are used as the new unseen domains to benchmark the performance.
In training, the classifier is trained using 60,000 images of MNIST. In order to generalizing new image phase, we use 10,000 images in this set to perturb and generalize new samples. To be convenient, all digit images are resized to $32 \times 32$ pixels. 
In testing, the proposed method is benchmarked on the testing set of MNIST and three other unseen digit datasets, i.e. USPS, SVHN and MNIST-M. The classification results using the proposed approach are compared against the pure LeNet classifier (Pure-CNN), and the Adversarial Data Augmentation (ADA) \cite{generalize-unseen-domain} methods. We also show the recognition results on these datasets using the Domain Adaptation methods, including: Adversarial Discriminative Domain Adaptation (ADDA) \cite{adda_cvpr2017}, Domain-Adversarial Training of Neural Networks (DANN) \cite{pmlr-v37-ganin15} and Image to Image Translation for Domain Adaptation (I2IAdapt) \cite{Murez_2018_CVPR}. It is notice Pure-CNN, ADA and our approaches do not require the target domain data during training. However, ADDA, DANN and I2IAdapt require the target domain data in the training steps.

The experimental results are shown in Table \ref{tab:performance_on_digit_classification}. The results on SVHN and MNIST-M shows that the proposed approach achieves better accuracy than ADA on these datasets. As we mentioned, our perturb phase generalizes images based on semantic space via the estimation of environment density. It helps our generated images are more diverse than the synthesized images using ADA method.  In USPS benchmarking, USPS and MNIST datasets have the similar environment conditions as shown in Figure \ref{fig:char_dbs}(A) and (C). Therefore, the image domain that USPS belongs to has been learned very well by the pure classifier and do not need any extra work for it. This scenarios eventually also happens to ADA method. The proposed method achieves better accuracy than the domain adaptation method, i.e. ADDA, on SVHN dataset.
However, these method requires images in new domains in training. Meanwhile our method do not required training images in new domains.

\begin{table}[t]
    \centering
    \caption{Experimental results on digit classification on four digit datasets. ADA and our proposed approaches \textbf{do not require} target domain data during training. ADDA, DANN, CoGAN and I2IAdapt \textbf{require} training data from target domains during training steps.}
    \begin{tabular}{|c|c|c|c|c|c|}
         \hline
         \textbf{Database} &  \textbf{MNIST} & \textbf{USPS} & \textbf{SVHN} & \textbf{MNIST-M} \\
         \hline
         Pure-CNN & 99.36\% & 82.46\% & 37.89\% & 56.93\% \\
         ADA        & 99.17\% & 81.77\% & 37.87\% & 60.02\% \\
         \textbf{Ours} & \textbf{99.17\%} & \textcolor{black}{\textbf{78.85\%}} & \textbf{40.22\%} & \textbf{60.51\%} \\ \hline \hline
         ADDA       & 99.29\% & 94.81\% & 32.20\% & 63.39\% \\
         DANN       & $-$    & $-$    &  $-$   & 76.66\% \\    
         CoGAN      & $-$    & 91.20\% &  $-$   & $-$ \\
         I2IAdapt   & $-$    & 92.10\% &  $-$   & $-$ \\
         \hline
    \end{tabular}
   
    \label{tab:performance_on_digit_classification}
\end{table}

\begin{figure*}[t]
	\centering \includegraphics[width=1.99\columnwidth]{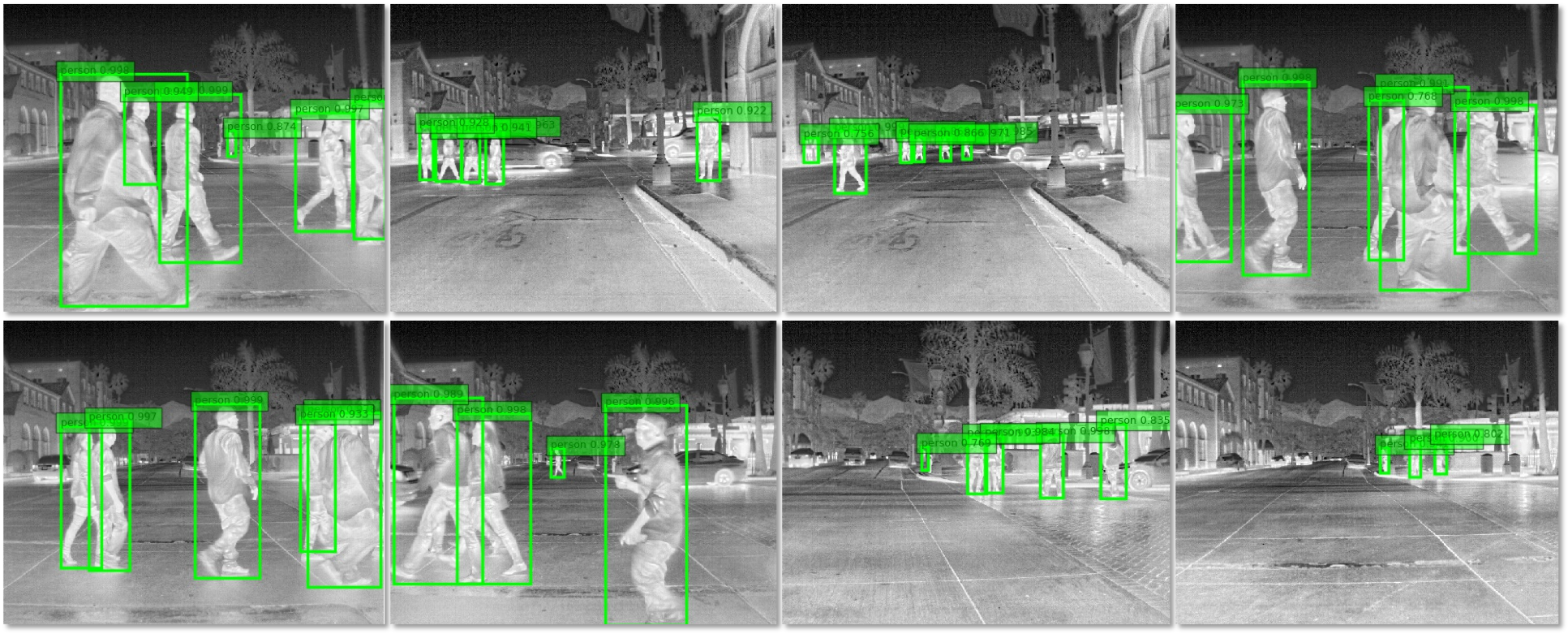}
	\caption{Examples of pedestrian detection results on thermal images using our proposed approach and Deformable-ConvNets training only on RGB images.}	
	\label{fig:thermal_det}
\end{figure*}

\subsection{Face Recognition on Unseen Domains}
\label{subsec:face}

In this experiment, our proposed approach is compared with Pure-CNN, ADA and ADDA methods in the face recognition application on three face recognition databases, including: Extended Yale-B, CMU-PIE and CMU-MPIE databases as shown in Figure \ref{fig:face_dbs}. In each database, the face images with normal lighting conditions will be selected as the source domain (Normal Illumination) and the face images with dark lighting conditions will be selected as the target domain (Dark Illumination). We use the same framework as Digit Recognition. All images are resized to $64 \times 64$ pixels. 
Table \ref{tab:performance_on_face_dataset} shows our experimental results on Extened Yale-B, CMU-PIE and CMU-MPIE datasets. The results show that our proposed method help to improve the recognition performance on new unseen domains where the lighting conditions are not known.

\begin{table*}[!t]
  \centering
  \caption{Experimental results of face recognition on Extended Yale-B \cite{yale_b_dataset}, CMU-PIE \cite{pie_dataset} and CMU-MPIE \cite{multi_pie_dataset} databases. It is notice that ADA and our proposed approaches \textbf{do not require} target domain data during training. ADDA \textbf{requires} training data from target domains during training steps. For Extended Yale-B, CMU-PIE and CMU-MPIE datasets, we split each dataset into two sets: training set (80\%) and testing set (20\%).}
  \begin{tabular}{|l|l|l|l|l|l|l|}
    \hline
    \multirow{2}{*}{\textbf{Database}} &
      \multicolumn{2}{c|}{\textbf{Extended Yale-B}} &
      \multicolumn{2}{c|}{\textbf{CMU-PIE}} &
      \multicolumn{2}{c|}{\textbf{CMU-MPIE}} \\ \cline{2-7}
                        & \textbf{Normal} & \textbf{Dark} & \textbf{Normal} & \textbf{Dark} & \textbf{Normal} & \textbf{Dark} \\
    \hline
    Pure-CNN & 98.90\% & 49.15\% & 96.09\% & 62.87\% & 99.95\% & 95.18\% \\
    ADA & 99.00\% & 53.08\% & 96.49\% & 62.69\% & 99.92\% & 96.08\% \\
    \textbf{Ours} & \textbf{99.73\%} & \textbf{70.09\%} & \textbf{97.32\%} & \textbf{67.87\%} & \textbf{99.83\%} & \textbf{98.25\%} \\ \hline \hline
    ADDA & 99.17\% & 75.28\% & 96.09\% & 70.33\% & 99.93\% & 97.71\% \\
    \hline
  \end{tabular}
  
    \label{tab:performance_on_face_dataset}
\end{table*}

\subsection{Pedestrian Recognition on Unseen Domains}
\label{subsec:ped}

This experiment aims for improving pedestrian detection on thermal images on the Thermal Dataset\footnote{\url{https://www.flir.com/oem/adas/adas-dataset-form/}}. This dataset includes both thermal and RGB images. We create two datasets for the pedestrian recognition: (1) RGB pedestrian and (2) Thermal pedestrian datasets. To make these datasets, pedestrian objects are cropped from images in the Thermal Dataset. Our model is trained only on RGB pedestrian dataset.
The baseline is trained on RGB pedestrian dataset and tested on Thermal pedestrian dataset. In the training phase, we use $5,000$ images to generalize new images, all images of two datasets are resized to $64 \times 64$ pixels. 
Table \ref{tab:pedestrian_recognition} shows our experimental results on RGB pedestrian dataset and Thermal pedestrian dataset. 

\begin{table}[]
    \centering
      \caption{Experimental results of pedestrian recognition on RGB pedestrian dataset and Thermal pedestrian dataset.}
    \begin{tabular}{|c|c|c|}
    \hline
\textbf{Database}   & \textbf{RGB}       & \textbf{Thermal} \\
    \hline
    Pure-CNN        & 96.61\%            & 94.44\%  \\
    ADA             & 98.05\%            & 95.42\%  \\
    \textbf{Ours}   & \textbf{97.04\%}   & \textbf{96.29\%}  \\
    \hline
    \end{tabular}
  
    \label{tab:pedestrian_recognition}
\end{table}

To further improve pedestrian detection, we apply our pedestrian recognition trained on RGB pedestrian dataset to the Deformable-ConvNets detector \cite{dai17dcn, dai16rfcn} trained on COCO \cite{coco_dataset} dataset. After the image proposal phase, we crop proposed bounding boxes and feed into our pedestrian recognition framework.

Table \ref{tab:pedestrian_detection} shows our experimental results of pedestrian detection on Thermal Dataset. Because of the abstract pedestrian shape (body shape) is keep on the thermal image. Therefore, the vanilla Deformable-ConvNets detector also have good results. Although our proposal helps to softly improve the results, the results prove that our method is promising.

\begin{table}[]
    \centering
     \caption{Experimental results of pedestrian detection on Thermal Dataset. The results are based on mean average precision (mAP) metric for object detection.}
    \begin{tabular}{|p{3cm}|c|c|}
        \hline
     \textbf{No. proposed bounding boxes}   &  \textbf{Detector}  & \textbf{Detector + Ours} \\
        \hline
        \multicolumn{1}{|c|}{300}           &        53.64\%      & \textbf{53.83\%}       \\
        \multicolumn{1}{|c|}{500}           &        53.21\%      & \textbf{53.54\%}       \\
        \multicolumn{1}{|c|}{700}           &        52.83\%      & \textbf{53.30\%}       \\
        \hline

    \end{tabular}
   
    \label{tab:pedestrian_detection}
\end{table}

\section{Conclusions}
\label{sec:concl}

This paper has introduced a new UNVP learning model approach that generalize well to different unseen domains. Only using training data from the source domain, we propose an iterative procedure that augments the dataset with examples from a fictitious target domain that is "hard" under the current model. On digit recognition, we benchmark on four popular digit recognition databases, i.e.  MNIST, MNIST-M, USPS, and SVHN. The method is also experimented on face recognition on Extended Yale-B, CMU-PIE and CMU-MPIE databases and compared against other the state-of-the-art methods.
In the problem of pedestrian detection tasks, we empirically observe that the proposed method learns models that improve performance across a priori unknown data distributions. 

{\small
\bibliographystyle{ieee}
\bibliography{egbib}
}

\end{document}